\documentclass[electronic]{vgtc}             

\ifpdf
  \pdfoutput=1\relax                   
  \usepackage{graphicx}                
  \DeclareGraphicsExtensions{.pdf,.png,.jpg,.jpeg} 
\else
  \usepackage{graphicx}                
  \DeclareGraphicsExtensions{.eps}     
\fi%

\graphicspath{{figures/}{pictures/}{images/}{./}} 

\usepackage{microtype}                 
\PassOptionsToPackage{warn}{textcomp}  
\usepackage{textcomp}                  
\usepackage{mathptmx}                  
\usepackage{times}                     
\usepackage{cite}                      
\usepackage{tabu}                      
\usepackage{multirow}
\usepackage{booktabs}                  
\usepackage{xcolor}
\usepackage{siunitx}
\usepackage{amsmath}
\usepackage{algorithm}
\usepackage{algpseudocode}
\algnewcommand{\LineComment}[1]{\State \(\triangleright\) #1}
\usepackage{soul}

\newcommand{\etal}{\textit{et al}. }

\newcommand{\norm}[1]{\left\lVert#1\right\rVert}

\makeatletter
\def\thanks#1{\protected@xdef\@thanks{\@thanks
        \protect\footnotetext{#1}}}
\makeatother

\newcommand\rurl[1]{%
  \href{http://#1}{\nolinkurl{#1}}%
}

\onlineid{7473}

\vgtccategory{Research}

\vgtcinsertpkg

\title{OA-SLAM: Leveraging Objects for Camera Relocalization in Visual SLAM}

\author{Matthieu Zins*
\and Gilles Simon* 
\and Marie-Odile Berger*
\thanks{*{\tt\small forename.name@inria.fr}. This work was supported by the MoveOn project (Inria - DFKI).}
}
        
\affiliation{\scriptsize Université de Lorraine, Inria, LORIA, CNRS}

\teaser{
  \centering
  \includegraphics[width=\linewidth]{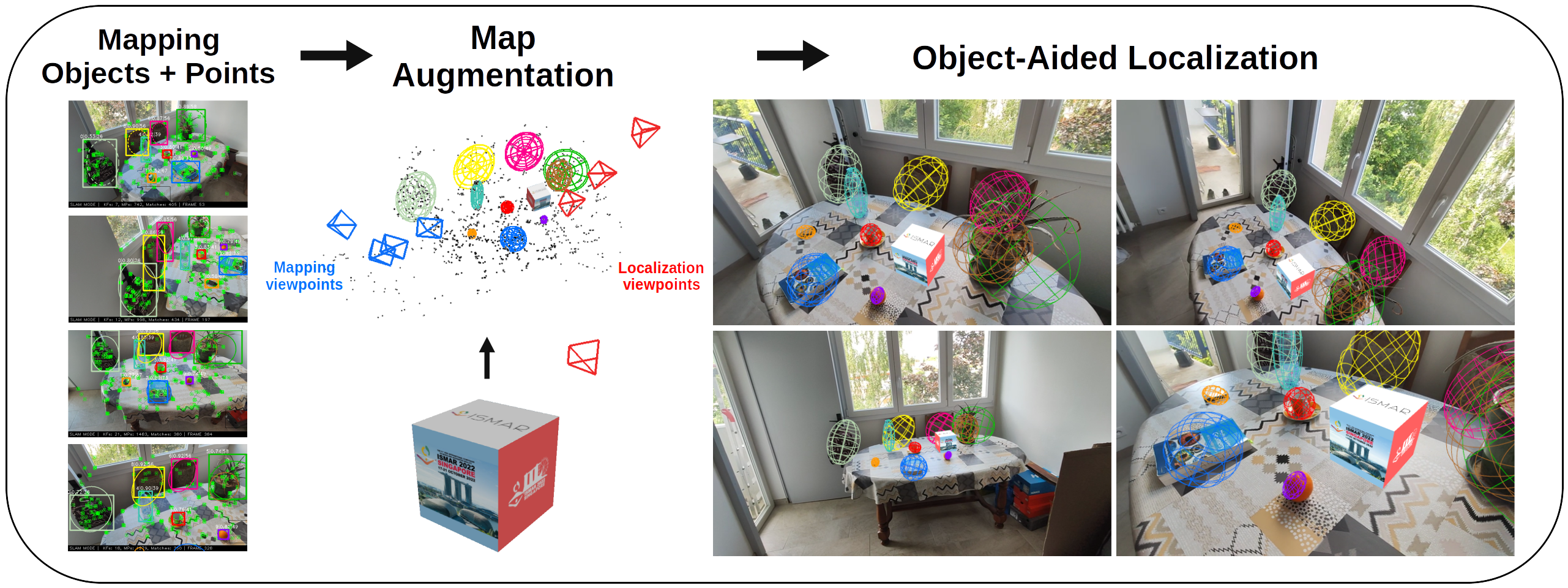}
  \caption{OA-SLAM enables on-the-fly object mapping and leverages them to improve the robustness of camera tracking reinitialization from a large variety of viewpoints. AR visualizations are shown on the right.}
  \label{fig:teaser}
}

\abstract{
In this work, we explore the use of objects in Simultaneous Localization and Mapping in unseen worlds and propose an object-aided system (OA-SLAM). More precisely, we show that, compared to low-level points, the major benefit of objects lies in their higher-level semantic and discriminating power. Points, on the contrary, have a better spatial localization accuracy than the generic coarse models used to represent objects (cuboid or ellipsoid).
We show that combining points and objects is of great interest to address the problem of camera pose recovery.
Our main contributions are: (1) we improve the relocalization ability of a SLAM system using high-level object landmarks; (2) we build an automatic system, capable of identifying, tracking and reconstructing objects with 3D ellipsoids; (3) we show that object-based localization can be used to reinitialize or resume camera tracking. Our fully automatic system allows on-the-fly object mapping and enhanced pose tracking recovery, which we think, can significantly benefit to the AR community.
Our experiments show that the camera can be relocalized from viewpoints where classical methods fail. We demonstrate that this localization allows a SLAM system to continue working despite a tracking loss, which can happen frequently with an uninitiated user. Our code and test data are released at \rurl{gitlab.inria.fr/tangram/oa-slam}.

} 

\CCScatlist{
\CCScatTwelve{Computing methodologies}{Computer graphics}{Graphics systems and interfaces}{Mixed / augmented reality};
\CCScatTwelve{Computing methodologies}{Artificial intelligence}{Computer vision}{Computer vision problems}
}

\begin{document}


\firstsection{Introduction}

\maketitle










Augmenting the real world with virtual information to create interactive experiences has gained attention with a large potential in entertainment, serious games, medical training, retail, tourism industry or maintenance of complex equipment. For example, museums and cultural institutions have become very fond of such applications, in which they see new ways of engaging their visitors.

In order to visually enrich the real world with virtual information, high-precision and robust camera pose tracking is crucial. A certain understanding of the scene is also necessary for positioning the virtual elements, when building object-level interactive applications. 

This is a well-known problem in computer vision and is called Simultaneous Localization and Mapping. It jointly builds a map of the environment and localizes the camera with respect to it. Many visual SLAM methods have been developed, turning RGB image sequences into sparse~\cite{ptam,dso,orbslam2}, semi-dense~\cite{svo,lsdslam}, or dense~\cite{dtam} maps. RGB-D cameras have also been successfully used in SLAM systems~\cite{badslam,3Dmapping,kinectfusion,bundlefusion}. However, they are limited to medium-sized indoor environments and struggle with highly reflective scenes, such as metallic structures. Also, depth sensors are not yet widespread enough to be used in consumer applications, whereas all the recent smartphones have a decent RGB camera.



Despite decades of research, using SLAM for Augmented Reality is still challenging.
Indeed, such applications are generally dedicated to the general public, including a majority of uninitiated users.
In that context,   fast  or abrupt camera motions often occur. In addition,  since the user is free of his/her motion, the system should  be able to restart from any location in the scene. These challenging conditions are, however,  generally  not supported by existing camera tracking systems.


Current state-of-the-art SLAM  methods, such as ORB-SLAM2~\cite{orbslam2}, rely on bag-of-words descriptors to find similar images and local appearance-based features, such as ORB or SIFT, to find matches between keypoints detected in the query image and landmarks in the map. However, using such low-level features is not well suited for relocalization when the viewing angle changes significantly, because of their limited invariance to viewpoint changes and because some surfaces used to build the map may no longer be visible.

On the contrary, impressive progress have been made in the field of object detection over the last few years and deep learning-based techniques are now able to detect objects from a large variety of viewpoints and environmental conditions with great robustness.
This naturally makes them good anchors to help visual-based camera localization.

In this work, we present a fully automatic object-aided (OA) SLAM system for unseen worlds, which is able to build a semantic map composed of 3D points and objects and leverages such high-level landmarks to improve its relocalization capability.
Our main contributions are:

\begin{enumerate}
    \item An improved relocalization method, combining the advantages of both objects and points, capable of estimating the camera pose from a large variety of viewpoints.
    \item A fully automatic SLAM system capable of identifying, tracking and reconstructing objects, on the fly.
    \item We demonstrate how our system can be used to reinitialize camera tracking on a previously built, and potentially augmented, map or to recover camera tracking after getting lost.
\end{enumerate}

\section{Related Work}
We place ourselves in the context of SLAM in unseen world and with minimal effort for deployment. Methods with precise 3D models of objects, such as SLAM++~\cite{slam++} and DeepSLAM++~\cite{deepslam++}, are thus not in the same line of work. For example, they require to have a database of CAD models of the objects seen during the deployment with specific networks training, which makes them painful to transplant into a new context.
To be able to consider unseen environments, we are here interested in object-based systems which use more general kinds of modeling: cuboids~\cite{cubeslam}, ellipsoids~\cite{quadricslam} or semantic point clouds~\cite{semanticfusion,meaningfull_maps}.

\subsection{Object Mapping}

Crocco proposed a closed-form formulation to estimate dual quadrics from multi-view object detections, using a simplified camera model in~\cite{crocco_sfm_with_objects}. Rubino then extended it to the pinhole camera model~\cite{rubino_ellipsoid_reconstruction}. In~\cite{forward_translation}, Chen~\etal addressed the problem of initial object estimation in the specific context of forward-translating camera movements, which commonly occurs in autonomous navigation. All these works are focused on  object mapping and assume that the camera poses and object associations and provided.

\subsection{Object-based Localization}

Weinzaepfel~\etal~\cite{planar-object-of-interest} proposed a method to compute the pose of the camera using dense 2D-3D correspondences between the objects present in a query image and those in reference images. However, this method is limited to planar objects.

More general objects, represented with ellipsoids, are used in~\cite{gaudilliere_ismar} and~\cite{ijcv}. However, compared to our work, these methods estimate the camera pose from objects only and assume a pre-built object map.
Also, \cite{gaudilliere_ismar} only estimates the position of the camera and the orientation is assumed to be known. For its part,~\cite{ijcv} is more focused on the improved 3D-aware elliptic detections of objects. They show how these better detections help to improve the accuracy of the estimated camera pose. However, the ellipse prediction network is trained on a specific scene, and thus, requires retraining in order to be used in a new scene, which is hardly acceptable for a SLAM system.

\subsection{Object-based SLAM}

A pioneering work, introducing objects in localization and mapping has been developed in~\cite{bao}, by Bao~\etal, in which they recognize and localize objects within a Structure-from-Motion framework.

McCormac~\etal~\cite{semanticfusion} and Sünderhauf~\etal~\cite{meaningfull_maps} fused RGB-D SLAM with semantic segmentation and object detection to obtain semantically annotated dense point clouds. In such works the semantic information is added to the map after its creation but does not inform localization. Also, the created maps are not object-centric but dense point clouds where each point carries a semantic label. 

In QuadricSLAM~\cite{quadricslam}, Nicholson~\etal derived a SLAM formulation which uses dual quadrics as 3D landmarks. They jointly estimate the camera pose and the dual quadrics parameters by combining odometry and high-level landmarks in factor graph-based SLAM. However, they assumed to have perfectly known data association, which considerably limits its application.
EAO-SLAM~\cite{eaoslam} integrated objects in a semi-dense SLAM and exploited different statistics to improve the robustness of data association. Objects are represented as cuboids or ellipsoids, depending of their nature and are assumed to be placed parallel with the ground. This requires to know the vertical direction of the scene.
Hosseinzadeh combined points, planes and quadrics into factor graph-based SLAM in~\cite{quadrics_and_planes_slam} and~\cite{realtime_monocular_hosseinzadeh}. They used an object detector network as well as a joint CNN to predict depth, surface normals and semantic segmentation, in order to capture the dominant structure of the scene and model point-plane, plane-plane and object-plane constraints. Their experiments show that incorporating objects and planes as new factors produce semantically meaningful maps and more accurate trajectories.
In SO-SLAM~\cite{so-slam}, Liao~\etal used manually extracted planes to add supporting constraints to objects, as well as, semantic scale priors and symmetry constraints. 

An object-SLAM specialized for autonomous navigation was presented in~\cite{high_speed}. They exploited bounding box detections, image texture, semantic knowledge and prior on objects shape (Toyota Camry) to infer ellipsoidal models and overcome the observability problem under forward-translating vehicle motions. However, only the object mapping part of their system is evaluated.


In CubeSLAM~\cite{cubeslam}, Yang~\etal used cuboids to represent objects. Their method is able to generate objects proposals from a single image using 2D bounding boxes and vanishing points sampling. The cuboids are then jointly optimized with camera poses and map landmarks. In~\cite{scale_drift}, Frost~\etal model objects with spheres and use them to resolve scale ambiguity and drift in SLAM.
Recently, category-specific deep shape priors have been exploited for object reconstruction within a real-time SLAM system~\cite{dsp_slam}.

\subsection{Object-based SLAM Relocalization}

While the previously described works integrate objects into SLAM systems, mainly as new factors in the global optimization, the relocalization problem is never discussed. Most systems are based on ORB-SLAM2, and thus, rely on the classical point-based approach. Leveraging objects for pose recovery is precisely the real novelty of our work. 
Only Dudek~\etal  ~\cite{dudek}  leverage semantic mapping in SLAM for relocalization. However, their method is more a post-processing step which is able to estimate the similarity transform between two object maps created from two sequences.

Very recently, Mahattansin~\etal improved relocalization in visual SLAM using object detections~\cite{Mahattansin}. Compared to our work, no object map is reconstructed and object detections are only used to better filter keyframes candidates. The camera pose is still estimated using point matches with the most similar keyframe.



\section{Motivations}

Most of the existing object-based SLAM systems~\cite{quadricslam,so-slam,high_speed,dsp_slam,realtime_monocular_hosseinzadeh,realtime_monocular_hosseinzadeh} integrated objects in the core of a SLAM system through a joint camera-landmark-object optimization. However,~\cite{so-slam} and the monocular version of~\cite{dsp_slam} noticed that it did not significantly improve the camera pose accuracy and~\cite{quadricslam} even observed a decreased accuracy compared to the point-based tracking of ORB-SLAM2. Objects represented with coarse models (cuboids or ellipsoids) might thus not be accurate enough to improve the camera pose tracking, but we nevertheless think that they are of great interest when the tracking gets lost.

Our fist motivation lies in the fact that current state-of-the-art object detectors reach a great robustness to viewpoint and illumination changes, which is very desirable for recovering the camera pose from a large variety of viewpoints.
Also, to our knowledge, this way of using objects has not been explored a lot in existing object-based SLAM systems. Only~\cite{dudek} and~\cite{Mahattansin} proposed a similar use of objects but they do not enable on-the-fly object map reconstruction and camera relocalisation.


 Our second motivation was the lack of a fully automatic system for building object-oriented maps. Indeed, most existing methods require external information or manual processing.
 For example, QuadricSLAM~\cite{quadricslam} and SO-SLAM~\cite{so-slam} assume that the problem of object association is solved. In their experiments, they manually annotated matches between the detected objects and the objects in the map. For their part, the authors of EAO-SLAM~\cite{eaoslam} assume that the objects are placed parallel with the ground, which necessitates to know the vertical direction of the scene. DSP-SLAM~\cite{dsp_slam} is able to provide fine object reconstructions, but at the cost of training one additional shape prior network for each class of objects encountered during the use of the system. In contrast, we chose to use a generic and relatively coarse object model, while privileging the fully automatic aspect of our system.
 







\section{Method}

\subsection{System Overview}

\begin{figure}[ht]
    \centering
    \includegraphics[width=\linewidth]{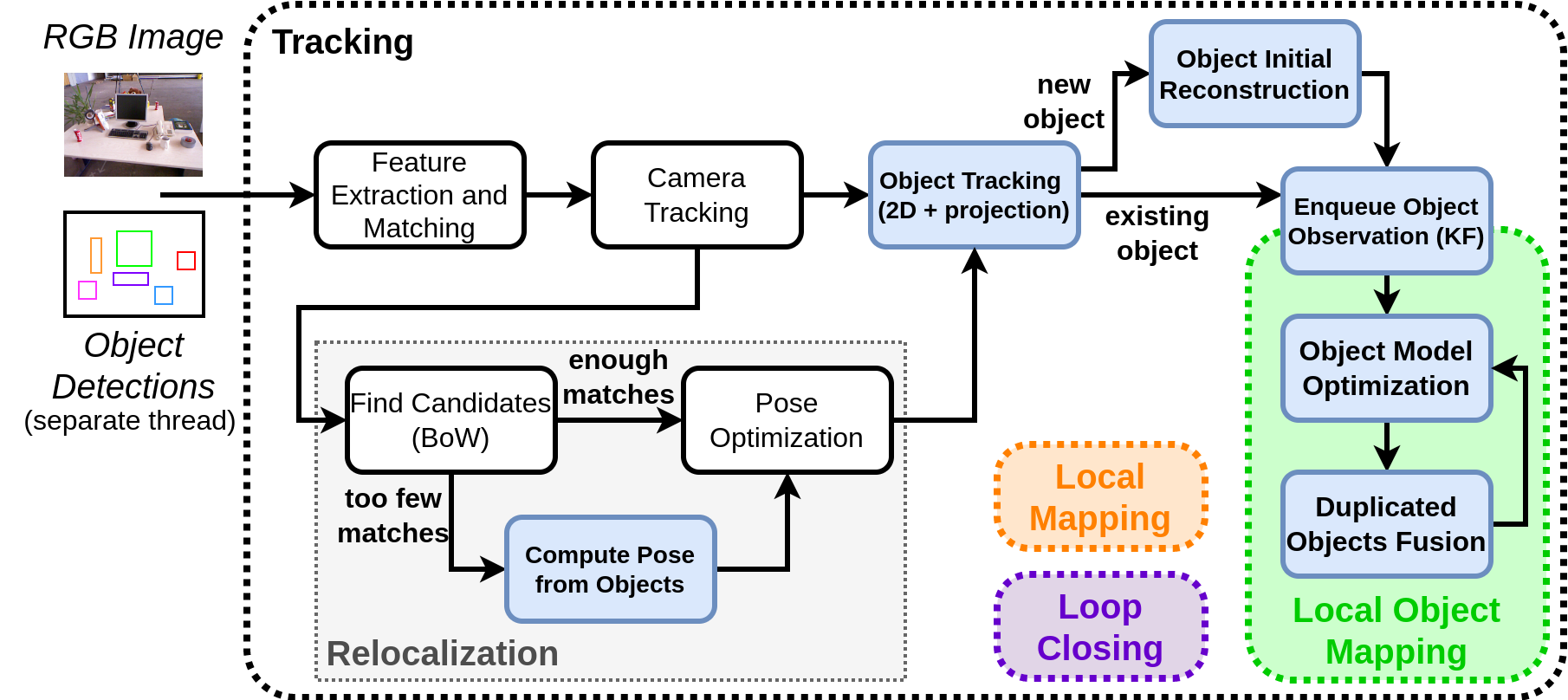}
    \caption{System: Blue items correspond to newly added elements within the ORB-SLAM2 backbone. Note that each module (Tracking, Local Mapping, Loop Closing and Local Object Mapping) is run in a separate thread.}
    \label{fig:system}
\end{figure}

Our system is detailed in Figure~\ref{fig:system}. It is based on ORB-SLAM2 (tracking, local mapping, loop closure) and augmented with additional modules dedicated to objects. These modules use the ellipse/ellipsoid modeling framework, described in section~\ref{subsec:object_model}, and follow the same strategy as points, i.e. objects are tracked over  frames, estimated in 3D, inserted into the map, and then, continuously refined.
In particular, object tracking and object initial reconstruction are added to the main tracking thread. They are respectively described in sections~\ref{subsec:data_association} and~\ref{subsec:initial_object_reconstruction}.
The local object mapping is handled in a similar way as the local point mapping and continuously refines the object models. It is run in a separate thread and is described in section~\ref{subsec:local_object_mapping}.
Finally, the relocalization module is enhanced by integrating objects, as described in section~\ref{subsec:reloc_using_objects}, greatly improving its robustness.

\subsection{Ellipsoidal Object Representation}
\label{subsec:object_model}
In this work, we model an object with an ellipsoid, in 3D, and its observation in images with an ellipse.
This is a coarse but lightweight representation which only requires nine parameters: three for its axes size, three for its orientation and three for its position. 
Also, an ellipsoid projects as an ellipse under any viewpoints, whose equation can be expressed in a closed-form manner using the dual space. In that space, the ellipsoid is defined by a $4\times 4$ matrix $Q^*$ and the ellipse by a $3\times 3$ matrix $C^*$, which are linked together through a projection matrix $P$ \cite{Hartley2004}:
\begin{equation}
    C^* = PQ^*P^T.
\label{equ:ellipsoid_projection}
\end{equation}
In comparison, it would not be possible with a cuboidal representation of an object. Indeed, matching the 3D box corners with the 2D box edges leads to high combinatorics.

\subsection{Object Detection and Association}
\label{subsec:data_association}

We use the state-of-the-art object detection network YOLO~\cite{yolov4} to obtain object detections in the video frames. Each detection includes an axis-aligned bounding box, a category and a detection score. For robustness, we only consider detections with a score higher than 0.5 and discard the others.

Establishing associations between object detections over time is a crucial part of our system. Given a set of detections in the current frame, the goal is to match each of them, either to an existing object track or decide to create a new one.  Associations are   firstly constrained  by the object categories. We  additionally take into account both  the overlap of detection boxes  as well as  points matching between boxes. This allows us  to  handle inaccurate  or partial object detections.


\subsubsection{Box-based Object Tracking}
\label{subsubsec:box_based_tracking}
Before being reconstructed, objects are tracked over frames in 2D, based on bounding boxes overlap and label consistency. This tracking is possible in the short term, when the motion between two frames is relatively small and smooth. However, it is prone to errors when detections are missing in some frames, when an object gets out of the camera field-of-view, or in case of abrupt camera motion.

Once 2D tracking of an object has been successfully performed on a sequence with a sufficient baseline, longer-term tracking can be obtained by considering its 3D reconstruction. For that, its ellipsoidal model is projected in the current frame (Eq.~\ref{equ:ellipsoid_projection}) and the overlap with the object detections in this frame is used to find associations.

In both cases, the optimal associations are found using the Hungarian algorithm~\cite{hungarian}, a well-known method for solving the assignment problem. This algorithm maximizes a total score of matching in order to find the best possible assignments between $N$ detections and $M$ objects. We define its score matrix at frame $t$ as


\begin{gather}
    \mathbf{S}^t = [s_{ij}^t]_{N\times M} \\
    s_{ij}^t =  \max(\mathrm{IoU}(\mathtt{D}_i^t, \mathtt{B}_j^{t-k}), \;
      \underbrace{\mathrm{IoU}(\mathtt{D}^t_i, \mathrm{box}(\mathrm{proj}(\mathtt{O}^t_j)))}_{\text{or 0 if reconstruction not yet available}}),
\end{gather}
where $\mathtt{D}_i^t$ is the i-th detection in the current frame, $\mathtt{B}_j^{t-k}$ is the latest bounding box associated to the j-th object and $\mathtt{O}^t_j$ is the j-th object ellipsoid. The operations 
\textit{IoU}, \textit{box} and \textit{proj} respectively stand for computing the intersection-over-union, the enclosing bounding-box and the projection in the current frame. The left IoU term refers to 2D bounding box tracking and the right one represents the long-term tracking by projection, only possible once an initial reconstruction of the object is available.



\subsubsection{Point-based Object Tracking}
\label{subsubsec:point_based_tracking}
The previously described long-term association method relies only on the geometry of the reconstructed ellipsoid, which is very interesting for small or texture-less objects, but can fail in case of truncated detections. Indeed, it is based on a global model of an object (ellipsoid), whose detection in the image may have some variance.
Therefore, we additionally leverage points which provide a more precise 2D localization.
During the camera pose estimation by the SLAM system, image keypoints are robustly matched to map landmarks. These matches can actually be used to link detections boxes and objects ellipsoids, using the following two statements:
\begin{itemize}
    \item In the image, a keypoint is linked to a detection if it is situated inside the bounding box.
    \item In the map, a point-landmark is linked to an object if it is situated inside its ellipsoid.
\end{itemize}

If, at least, $\tau$ of such point-based matches exist, between a detection and an object, the association is considered. In our experiments, we used $\tau=10$. This tracking approach is only enabled once the object has been reconstructed and is particularly useful for highly-textured objects or when the geometry-based tracking fails.

\bigbreak

\subsection{Initial Object Reconstruction}
\label{subsec:initial_object_reconstruction}

Our main idea is to generate 3D object hypotheses  quickly after detection, in order that model based tracking can be performed in addition to box-based tracking. The validation and the integration of the object in the map comes later on,  if tracking in subsequent frames  is in coherence with the initial hypothesis. Otherwise, the object  hypothesis is rejected.

Once an object has been successfully tracked over frames with a sufficient change in viewing angles, our system creates an initial rough estimate of its 3D ellipsoid. Before reconstructing, we check that the maximum angle between the rays passing through the camera center and the center of the object detections  is above~\SI{10}{\degree}.
Rubino~\etal proposed a method for reconstructing an ellipsoid from elliptic detections in images, however, we observed numerical instability for small baselines. In order to obtain a 3D estimate of an object as soon as possible, we opted for a simpler but more reliable approach, where an object is initially reconstructed as a sphere, and then, refined in the form of an ellipsoid as more views arrive. The position of this sphere is triangulated from the centers of the bounding boxes and its radius is determined as the mean bounding box size, back-projected at the estimated position, such that
\begin{equation}
    radius = \frac{1}{2n} \sum_{i=1}^{n} \frac{z_i}{2} \left(\frac{w_i}{f_x} + \frac{h_i}{f_y}\right),
\end{equation}
where $z_i$ is the z-coordinate of the object center in the coordinate system of the i-th camera, $w_i$ and $h_i$ are the width and height of the detection box in the i-th frame, $f_x$ and $f_y$ are the camera intrinsics and $n$ is the number of frames in which the object has been tracked.

The sphere is then refined as an ellipsoid (its orientation is set to identity) and its axes and position are updated when the object is detected in new images. This refinement is expressed in the form of a minimization of reprojection errors, similarly to the optimization described in the next section.
Note that, from here, the data association can leverage this 3D estimate to search possible matches by projection.
Once the object has been reconstructed and refined in a sufficient number of frames (typically 40 frames), the  object is integrated in the map provided that the overlap between its projections and detections in all the frames where the object was tracked is sufficient. We set the minimal overlap threshold at 0.3.

It should be noted that, although  we use relatively low thresholds to favor object tracking, our reconstruction  method is robust both to false object detection and to false associations. Indeed,  false-positive detections that YOLO might generate are generally not stable enough over time to be tracked. In addition,  the requirement of a minimal overlap  in  all the frames in which an object is detected ensures that only coherent objects are integrated in the map.



\subsection{Local Object Mapping}
\label{subsec:local_object_mapping}

\subsubsection{Objects Refinement}

Similarly to point-landmarks in the local bundle adjustment of ORB-SLAM2, object models are also regularly refined. Each time a new keyframe observes an object present in the map, this object is updated through the minimization of a reprojection error.
We use the Wasserstein distance between the ellipse inscribed inside the detection box and the projection of the estimated ellipsoid as objective function. This distance has already been used as cost between ellipses to train an ellipse predicting neural network in~\cite{knots_detection}. In this distance, ellipses are interpreted as 2D Gaussian distributions $\mathcal{N}(\mu,\,\Sigma)$ such that

\begin{equation}
        \mu = \left[\begin{array}{c}
            c_x \\
            c_y 
        \end{array}\right], \;\;
        \Sigma^{-1} = R(\theta) \left[\begin{array}{cc}
            \frac{1}{\alpha^2} & 0 \\
            0 & \frac{1}{\beta^2}
        \end{array}\right] R(\theta)^T,
\end{equation}
where $[c_x, c_y]$ are the ellipse the center, $[\alpha, \beta]$ are its axes length and $\theta$ is its orientation. 
The Wasserstein distance between two Gaussian distributions $\mathcal{N}_1(\mu_1,\,\Sigma_1)$ and $\mathcal{N}_2(\mu_2,\,\Sigma_2)$ is computed by
\begin{equation}
\label{equ:Wasserstein}
\begin{split}
    \mathcal{W}_2^2(\mathcal{N}_1, \mathcal{N}_2) = & \norm{\mu_1-\mu_2}_2^2 \\
    & + Tr(\Sigma_1 + \Sigma_2 - 2 (\Sigma_1^{\frac{1}{2}} \Sigma_2 \Sigma_1^{\frac{1}{2}})^{\frac{1}{2}}).
\end{split}
\end{equation}
The ellipsoidal shape of the i-th object is thus refined by minimizing
\begin{equation}
\label{eq:object_refinement}
    \sum_{j=0}^{N} \sigma_j^{-1} \mathcal{W}_2^2(E_{ij}, P_j Q_i^* P_j^T),
\end{equation}
where $E_{ij}$ is the ellipse inscribed in the j-th detection, $Q_i^*$ is the dual matrix of the i-th object, $P_j$ is the projection matrix of the j-th keyframe, $\sigma_j$ is the score of the detection in the j-th keyframe and $N$ is the number of observations of the object.




\subsubsection{Objects Fusion}

In some cases, an object might be duplicated in the map. This can happen when a detected object was not visible for a few frames and data association fails to correctly re-match it with the existing track and inserts a new object in the map. To prevent such cases, our system regularly checks for duplicated objects. 
Two objects of the same category are considered as a unique object if their 3D aligned boxes IoU, which is very fast to compute, exceeds a certain threshold (0.2 in ours experiments), if the center of one ellipsoid lies inside the other one, or if they share more than $\tau$ common 3D landmarks. 
In such a case, the detection boxes tracked for both objects in keyframes are combined and a new ellipsoid is initialized, following the procedure described in section~\ref{subsec:initial_object_reconstruction}, but only on keyframes.

\subsection{Relocalization using Objects}
\label{subsec:reloc_using_objects}

The original relocalization method of ORB-SLAM2 offers a good reliability but often fails when the query image is far from the past camera trajectory.  Indeed, it uses Bag-of-Word descriptors to find similar keyframe candidates and searches for point matches, which fails frequently when the viewpoint on the reconstructed map differs significantly from the keyframe. As described in Figure~\ref{fig:system}, we enhance the relocalization with an object-based approach, more robust to viewpoint changes, which is triggered when too few point correspondences have been established. Indeed, objects learnt from large database have the advantage that they can be detected from a large variety of viewpoints (front, back, top, side, ...), thus opening the way towards relocalization from any position  without specific knowledge of the objects in the scene.

Knowing that poses computed from Perspective-n-Point (PnP) are more accurate than those obtained from object correspondences, our main idea is to guide point matching with the pose computed from 2D/3D object correspondences.   This pose  is generally sufficient so that projection of the landmark points  are similar to detected points, thus allowing to  perform easily point correspondences and to use PnP for localization.
 
 Our object-based approach is a modified version of the algorithm presented in~\cite{ijcv}, which jointly determines object correspondences between the query image and the map and estimates the pose of the camera.
Pairs of ellipse-ellipsoid are established based on their categories. At each iteration, a minimal set of three pairs are selected and the camera pose is computed with the Perspective-3-Point (P3P) algorithm on the ellipses and ellipsoids centers. P3P provides four potential solutions.
For each pose, ellipsoids are projected and associated to detections based on their overlap.
A cost is calculated as the sum of $1-IoU$ for each associated pairs and the pose with the minimal cost among the four P3P solutions is selected. These poses are used to identify keypoint-landmark correspondences through the local matching step  of ORB-SLAM2.
Finally, the pose with the minimal cost and with more than 30 keypoint-landmark matches is selected and refined on points. The SLAM system can then resume tracking.
If keypoint-landmark correspondences could not be established, for example in a texture-less environment, the pose with the minimal cost is selected, without refinement. In that case, the current camera pose is estimated, but tracking can not be recovered and the system will trigger a new relocalization procedure when the next image arrives.

%








\section{Experiments and Results}\label{sec:results}

\subsection{Scenes and Objects}

To evaluate our approach, we used two scenes from TUM RGB-D dataset~\cite{tum_rgbd}: \textit{fr2/desk} and \textit{fr3/long\_office\_household}. This dataset provides ground truth poses for the camera trajectory. However, it is limited to one scan per scene with a mostly orbital camera trajectory.

We thus also recorded our own sequences using a standard smartphone camera. This allowed us to evaluate our method in more diverse environments (Fig.~\ref{fig:eval_localization_3_scenes}) and from a larger variety of viewpoints, in terms of both angle and scale (Fig.~\ref{fig:eval_reloc_museum_traj}).
We used COLMAP~\cite{colmap} to obtain ground truth camera poses.

It also allowed us to use both very general objects (book, chair, cup, ...), but also more specific ones (statue, amphora, ...). The \textit{sink} and \textit{desk} scenes in Figure~\ref{fig:eval_localization_3_scenes} show, for example, how our system  can be  used in an everyday life environment, using an off-the-shelf object detector. More specific objects, in particular texture-less statues, have also been tested (see Figures~\ref{fig:eval_reloc_museum_traj} and~\ref{fig:eval_localization_3_scenes}). This gives a good insight of how our system can be used for AR applications in museums, for example. For these objects, YOLO has been fine-tuned on a few manually annotated images (around 50 images).

The full videos of the experiments are available in the accompanying video.




\subsection{Object Mapping}



\begin{figure}[ht]
    \centering
    \includegraphics[width=\linewidth]{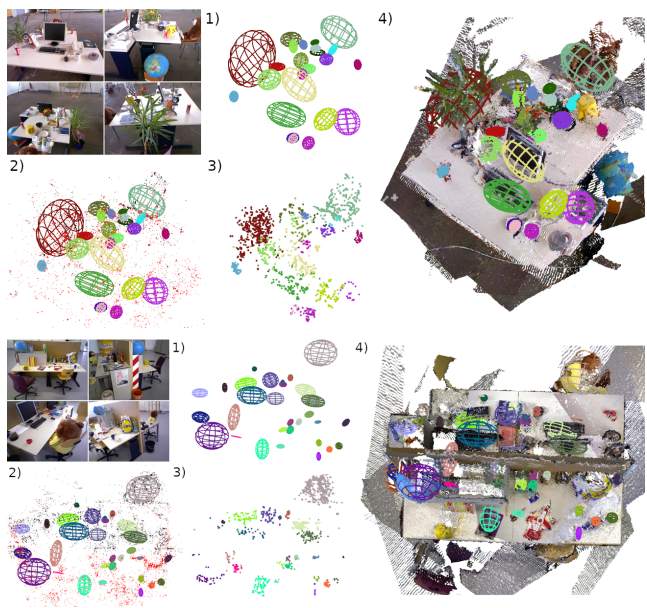}
    \caption{Obtained map on \textit{fr2/desk} (top) and \textit{fr3/long\_office\_household} (bottom). Upper-left images give an overview of the video sequences. 1)~Object map; 2) Object and point map; 3) Points associated to objects; 4) Object map displayed over a dense reconstruction of the scene (only for illustration).}
    \label{fig:eval_mapping_fr2_fr3}
\end{figure}

In this section, we first evaluate the object mapping part our system, which builds ellipsoidal models of the objects in the scene, on the fly, using the computed camera poses and object detections. Figure~\ref{fig:eval_mapping_fr2_fr3} shows the reconstructed map on \textit{fr2/desk} and \textit{fr3/long\_office\_household} sequences. We can observe that our coarse object models are globally well positioned on the objects in the map.

\begin{figure}[ht]
    \centering
    \includegraphics[width=\linewidth]{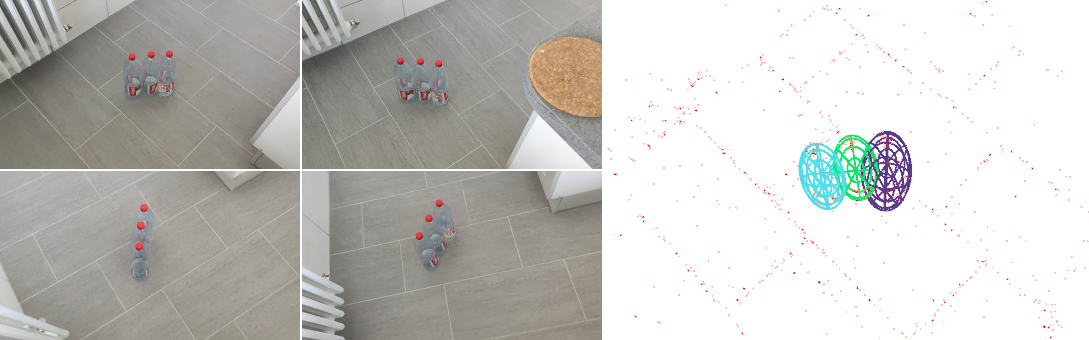}
    \caption{Obtained map for duplicated objects placed side-by-side. The images on the left give an overview of the sequence.}
    \label{fig:eval_mapping_bottles_1}
\end{figure}

Figure~\ref{fig:eval_mapping_bottles_1} shows a small but challenging scene, where three duplicated objects are placed side by side. Indeed, object classes can not be used to constraint data association and the objects are occluded when seen from the side. Our system still manages to build three accurate ellipsoidal models.

\begin{figure}[ht]
    \centering
    \includegraphics[width=\linewidth]{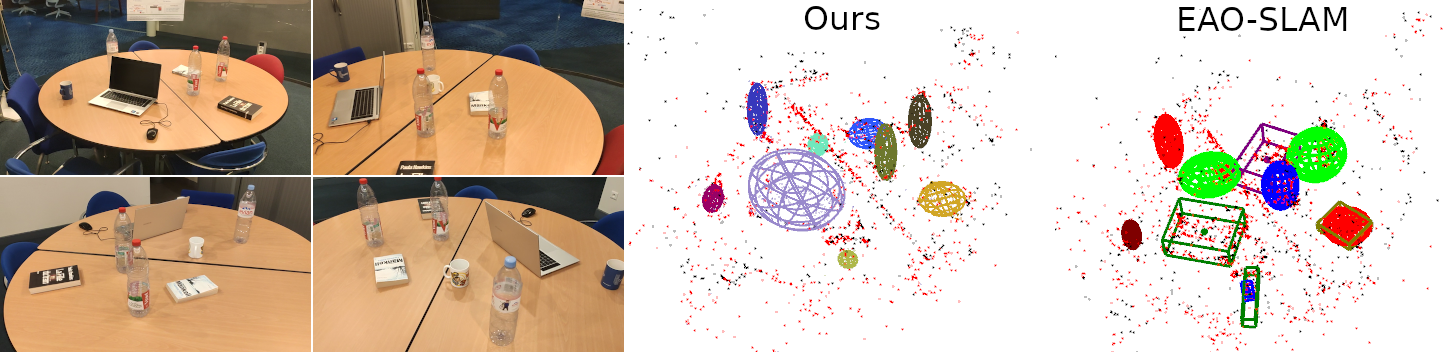}
    \caption{Comparison of the obtained maps using our method vs. EAO-SLAM~\cite{eaoslam}. The semi-dense map created by EAO-SLAM is not shown here, as they only use it for visualization.}
    \label{fig:eval_mapping_bulle_1}
\end{figure}

Finally, we compare our reconstructed map with the one built with EAO-SLAM \cite{eaoslam}  in Figure~\ref{fig:eval_mapping_bulle_1}. We used the code they released at: \url{ https://github.com/yanmin-wu/EAO-SLAM}. The scenario is simple with objects placed on a table and an orbiting camera trajectory. The same object detections were passed to both methods. For EAO-SLAM, we had to manually set the vertical direction so that objects can be estimated parallel to the ground. Both methods are based on ORB-SLAM2 and reconstruct similar sparse point clouds. In terms of objects, our method reconstructs precise ellipsoidal models of all the objects present on the table (the relatively large laptop, the thin bottles and the small mouse).
EAO-SLAM makes a distinction between objects with a regular or non-regular shape and represents them respectively with cuboids and ellipsoids. 
We can observe two cuboids created over the mouse and the black book (the red ellipsoid on the right side of the map) which should not exist. The bottles are represented with thicker ellipsoids which fit less well their shape. Finally, the mug placed behind the laptop has a much larger ellipsoidal reconstruction (the green ellipsoid in the middle in EAO's map and the small light-blue ellipsoid in our map).






\subsection{Objects versus Points}

\subsubsection{Relocalization}
In this section, we analyze the benefits of combining objects and points for relocalization.
The scenario of the experiment is the following: we first map the scene from limited viewpoints, mostly from one side, with our SLAM system and then call the relocalization procedure on query images from different viewpoints. For \textit{fr2/desk}, we split the sequence in two parts: the first 700 frames were used for mapping while the rest is used for relocalization. We also recorded our own sequences enabling more diverse viewpoints, in terms of angle and scale.

\begin{figure}[ht]
    \centering
    \includegraphics[width=\linewidth]{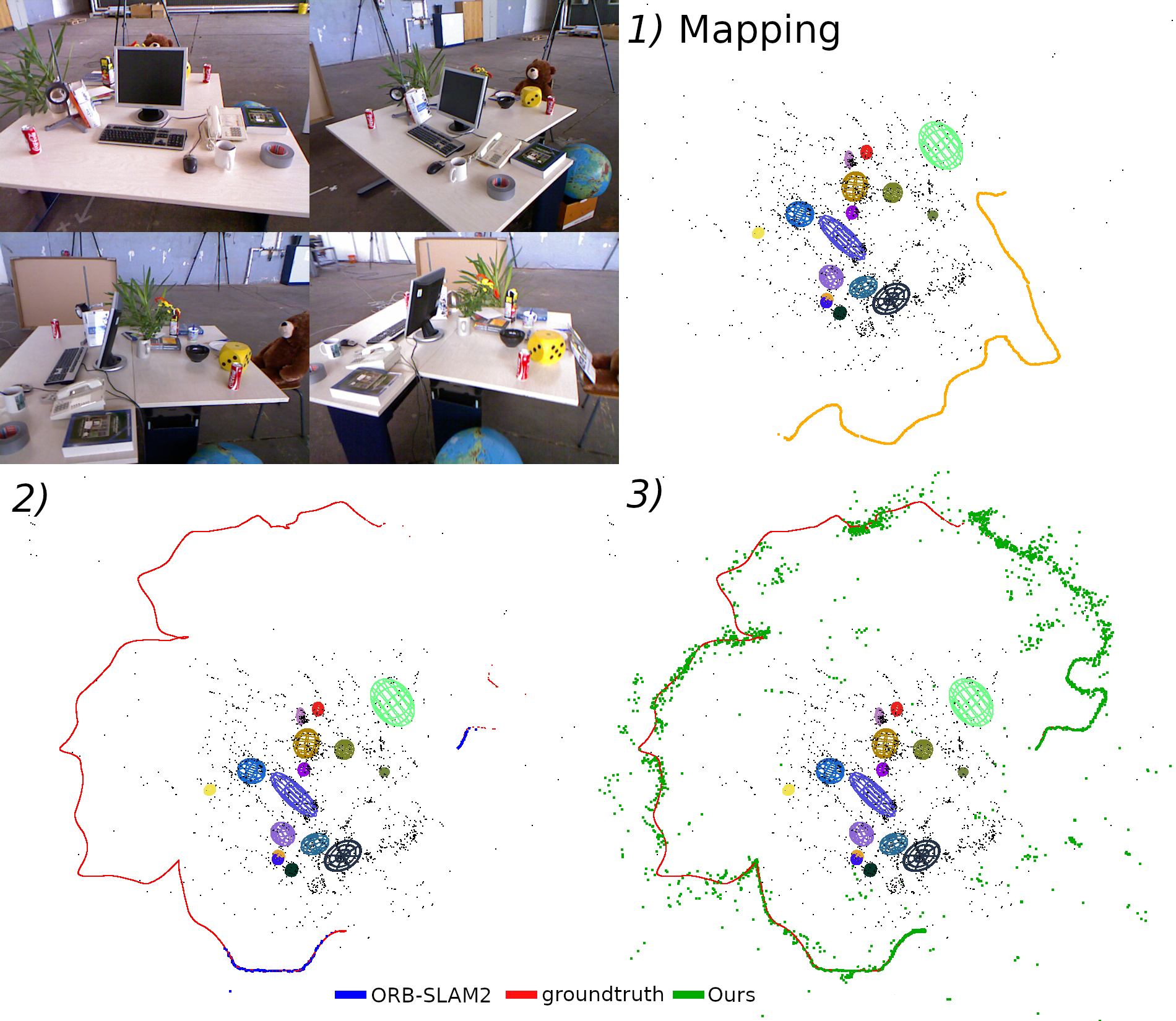}
    \caption{Estimated camera positions by the relocalization module (frame-by-frame) on \textit{fr2/desk}. The top-left images give an overview of the frames used for mapping. 1) The orange camera trajectory was used for mapping; 2) Relocalized camera using ORB-SLAM2 in blue; 3) Relocalized camera using our system in green. The ground truth trajectory is in red.}
    \label{fig:eval_reloc_fr2_desk_traj}
\end{figure}

\begin{figure}[ht]
    \centering
    \includegraphics[width=\linewidth]{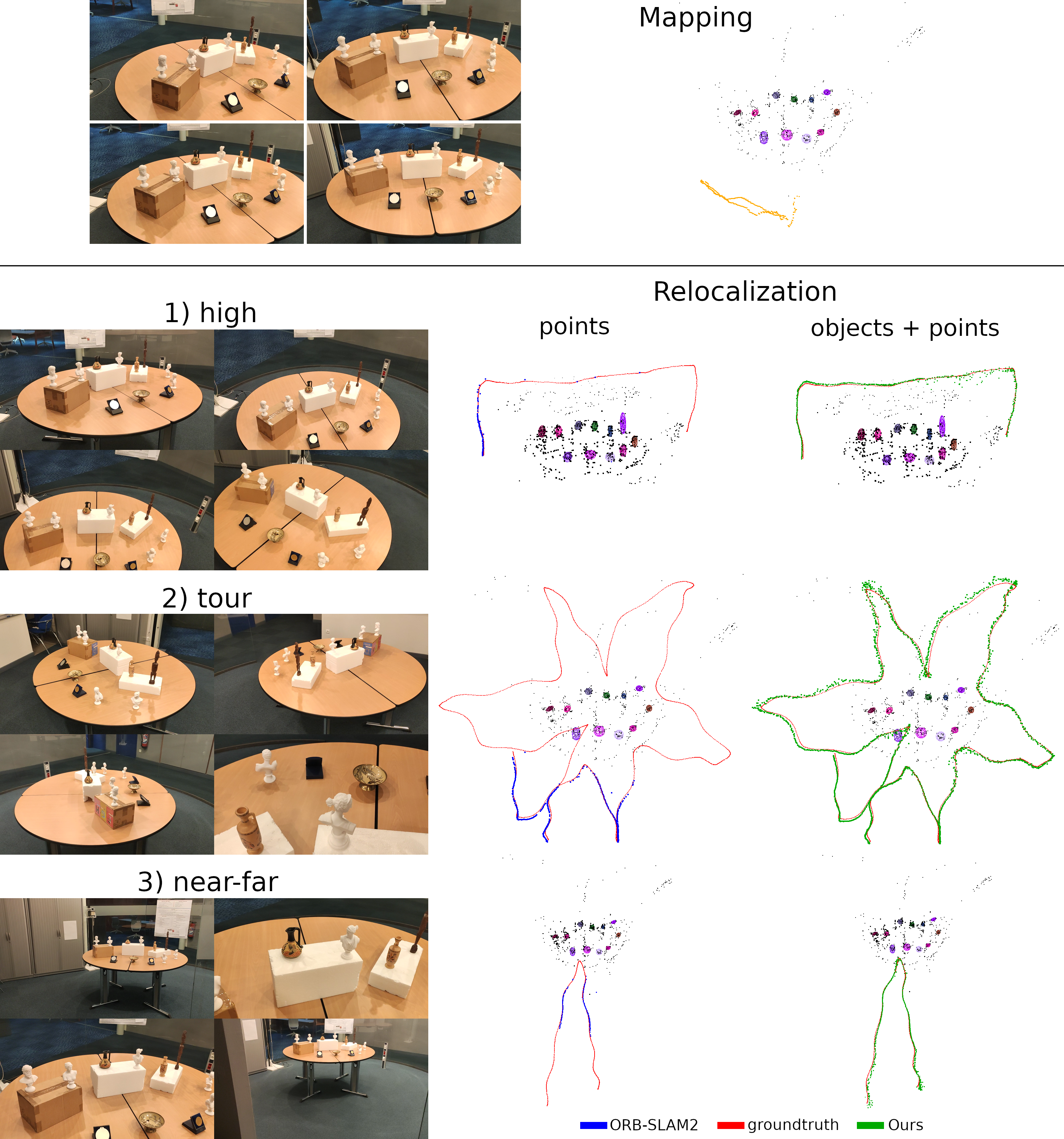}
    \caption{Estimated camera positions by the relocalization module (frame-by-frame) on a custom scene with a large viewpoints variety. Top: an overview of frames used for mapping and the obtained map with the estimated camera trajectory in orange.
    Bottom: relocalization results obtained on 3 test sequences. The images give an overview of the query frames. Camera positions estimated with ORB-SLAM2 are in blue, with our method in green and the groundtruth in red.}
    \label{fig:eval_reloc_museum_traj}
\end{figure}

\begin{figure}[ht]
    \centering
    \includegraphics[width=\linewidth]{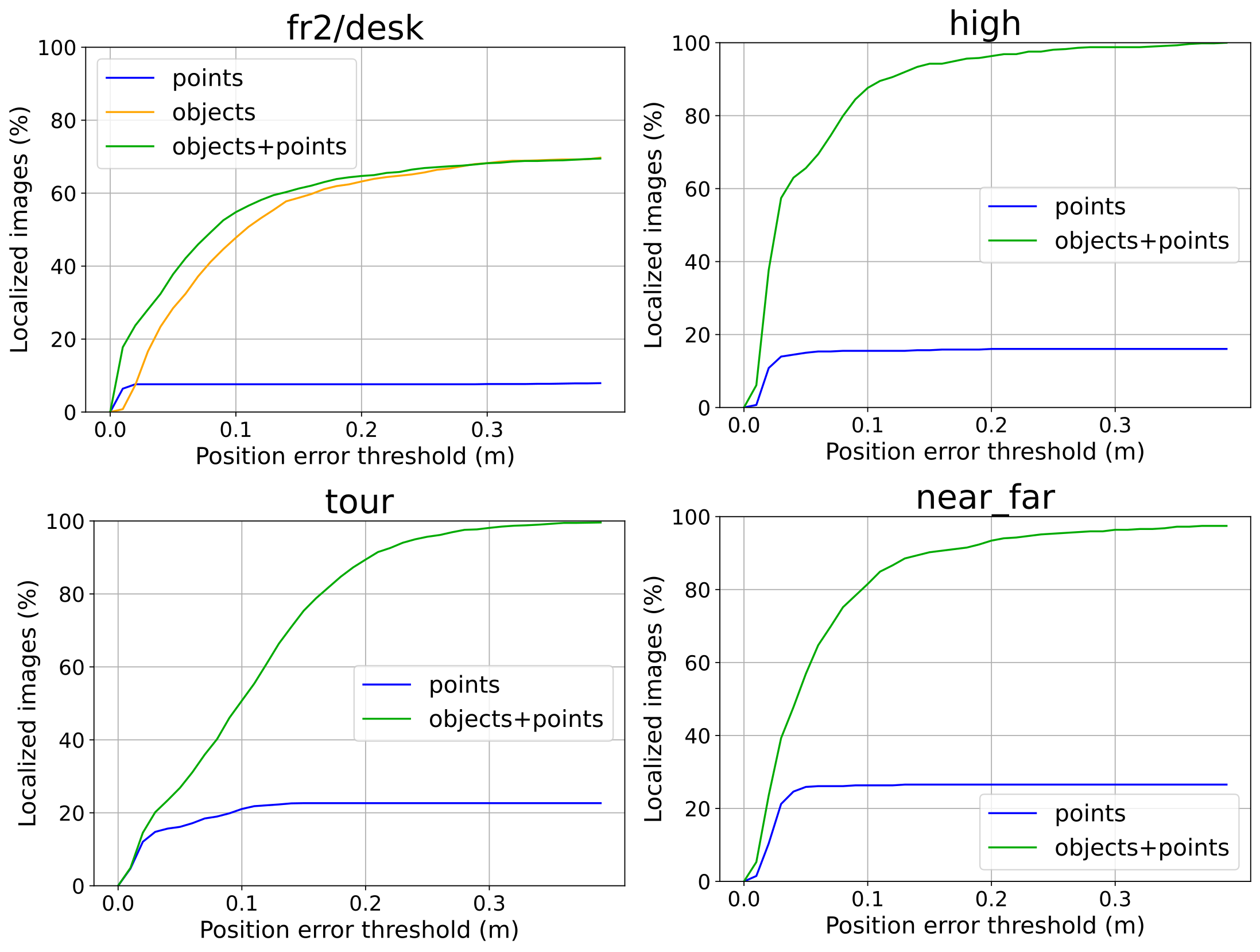}
    \caption{Percentage of estimated camera positions with respect to an error threshold on the position.}
    \label{fig:eval_reloc_curves}
\end{figure}

Figures~\ref{fig:eval_reloc_fr2_desk_traj} and \ref{fig:eval_reloc_museum_traj} compare our estimated camera positions with the results obtained using the classical point-based method available in ORB-SLAM2. We can clearly see that our method is able to accurately localize the camera all around the scene. This is achieved thanks to the robustness of object detectors, which can detect objects even from behind. The \textit{tour} and \textit{near-far} test sequences in Figure~\ref{fig:eval_reloc_museum_traj} significantly varies the distance from the scene. The accuracy of the estimated poses only slightly decrease for query images taken very far or very close to the scene. On the contrary, the original point-based relocalization is only able to produce pose estimates for viewpoints close to the ones used for mapping. This is also visible on the curves in Figure~\ref{fig:eval_reloc_curves}, which show that a much larger proportion of images can be localized. Our method does not improve the accuracy of the point-based approach, mainly due to the coarse object representation, but significantly enlarges its operating area. The plateau at around 70\% of localized images with our method on \textit{fr2/desk} can be explained by the fact that no or only one object is visible in a part of the frames. We also evaluated an object-only approach on \textit{fr2/desk}. The results in Figure~\ref{fig:eval_reloc_curves} show that it keeps the robustness of objects but loses the accuracy of points (see the bottom left part of the curve). 

\subsubsection{Integrating Objects in Bundle Adjustment}
\label{subsec:exp_objects_in_ba}

\begin{figure}[ht]
    \centering
    \includegraphics[width=\linewidth]{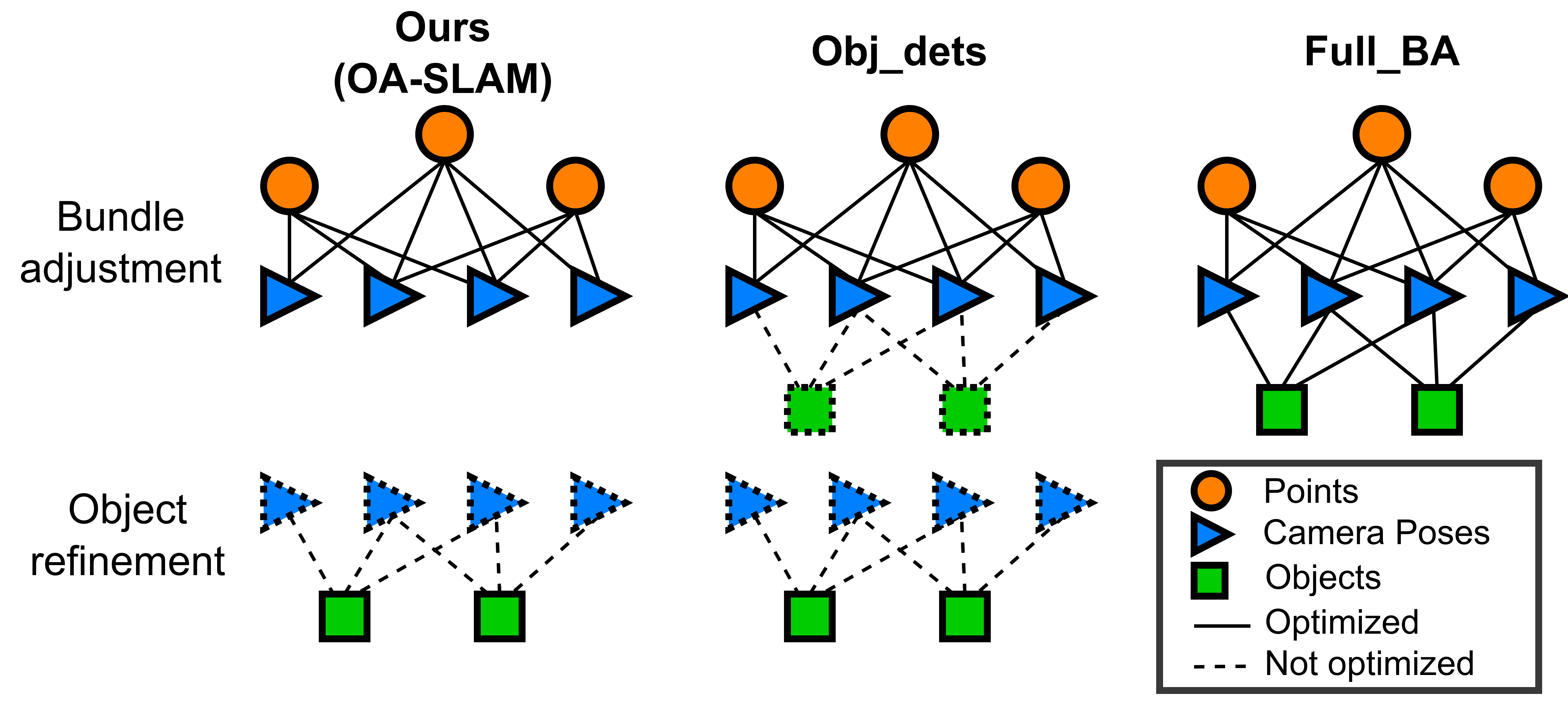}
    \caption{Three configurations for integrating objects in the SLAM bundle adjustment.}
    \label{fig:objects_in_ba_3_versions}
\end{figure}

In the previous experiments, we showed how objects are reconstructed by our SLAM system and why reasoning about such higher-level landmarks is beneficial for relocalization. In this section, we explore the question of using objects during camera tracking. We recall that, in our system, objects are not involved in camera tracking. They are refined independently in a separate optimization and are used for relocalization only. We thus created two other versions involving objects in the bundle adjustment (see Figure~\ref{fig:objects_in_ba_3_versions}). A first one, called \textit{Obj\_dets}, where objects are integrated in the bundle adjustment without updating their ellipsoidal models.
And a second one, called \textit{Full\_BA}, in which the objects models are completely integrated inside the bundle adjustment, together with camera poses and point-landmarks. The difficulty of combining point-based and object-based factors is that their cost need to be balanced. We thus arbitrarily scaled the object residuals in order to have values of the same order of magnitude as point residuals.
 
We measured the error of the estimated positions of keyframes on the fr2/desk and fr3/long\_office\_household sequences for different configurations of the bundle adjustment.
The results, available in Table~\ref{tab:objects_in_ba},  do not show a clear improvement by integrating objects in the graph-based optimization, and even decrease the pose accuracy in the case of the full camera-point-object bundle adjustment. Note, that we did not compare with ORB-SLAM2 because, in such scenarios of pure camera pose tracking, it is equivalent to our method, using the first configuration.

 We believe that coarse object models (ellipsoids) combined with simplified detections (axis-aligned bounding boxes), are not able to provide the same level of accuracy as current state-of-the-art point-based tracking in sufficiently textured areas. Similar results were notably obtained by ~\cite{quadricslam} and ~\cite{so-slam}, whereas~\cite{realtime_monocular_hosseinzadeh} noted slight improvements.

\begin{table}[ht]
  \scriptsize%
	\centering%
  \begin{tabu}{%
	r%
	*{7}{c}%
	*{2}{r}%
	}
  \toprule
     \multirow{2}{*}{Sequence} & OA-SLAM & \multirow{2}{*}{Obj\_dets} & \multirow{2}{*}{Full\_BA} \\
    & (ours) & & \\
  \midrule
	fr2/desk  & \textbf{0.808} &  0.901 &  0.860  \\
    fr3/long\_office\_household & \textbf{1.180} & 1.978 & 2.143 \\
  \bottomrule
  \end{tabu}%
  \caption{RMSE (\SI{}{cm}) of keyframes Absolute Trajectory Error (median value obtained on 20 runs).}
  \label{tab:objects_in_ba}
\end{table}

\subsection{Time Analysis}

The speed of our system depends on the number of objects present in the scene and their diversity of category. The tracking part takes a median time of \SI{27}{\ms} per frame, only slightly more than ORB-SLAM2 which takes \SI{25}{\ms}. Our object-aided relocalization takes a median time of \SI{22}{\ms} per frame.
We measured these times on the scene shown in Figure~\ref{fig:eval_reloc_museum_traj} using an Intel Xeon 3.6GHz CPU.  These durations are totally satisfactory for the targeted applications. Also, note that this scene includes a relatively high number of objects (11), with in particular, five objects of the same type (the statues) which increases the number of potential object associations.



\subsection{Application to AR}
We demonstrate here two direct applications of the improved relocalization capabilities. Scenes of everyday life, in a kitchen or in a dining room, are considered in this section. These scenes are complex since the notion of objects is less obvious than in the previous examples, especially for the sink. The term "object" is used in a broad sense. It refers to any element of the scene that could be detected using a specifically trained object detector network.

\subsubsection{Reinitializing 3D Tracking}

\begin{figure*}[ht]
    \centering
    \includegraphics[width=\linewidth]{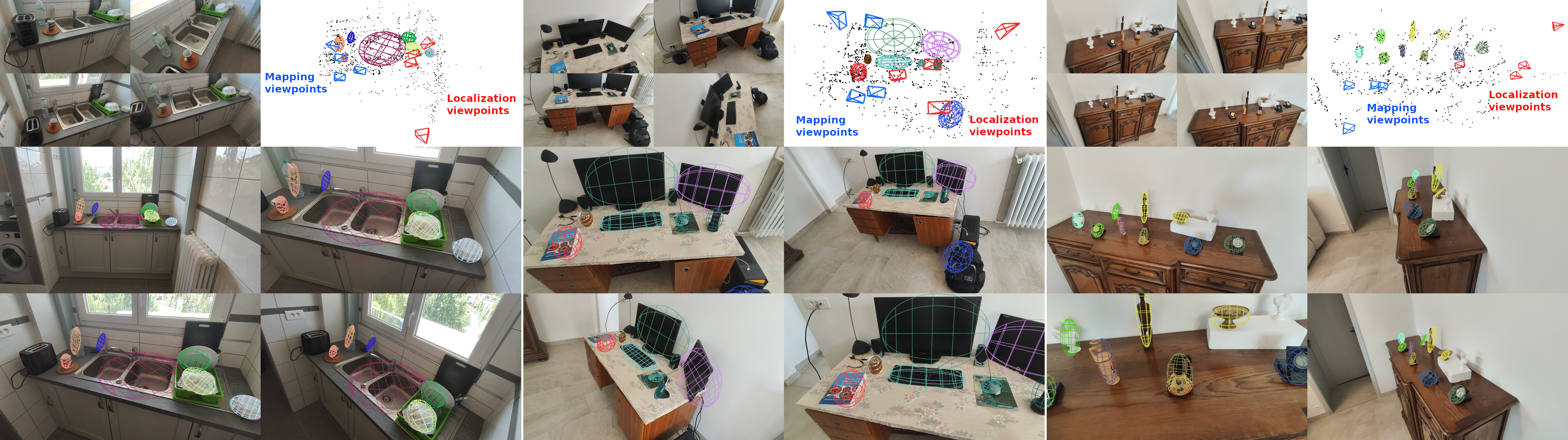}
    \caption{Camera tracking reinitialization on three scenes (\textit{sink}, \textit{desk}, \textit{museum}). For each scene, an overview of the frames used for mapping are shown in the top-left corner and the four frames below correspond to query images localized with our system.}
    \label{fig:eval_localization_3_scenes}
\end{figure*}

The enhanced relocalization capability is particularly interesting to initialize camera 3D tracking in AR applications. Once a map of the working area has been built, and potentially augmented with virtual elements, the camera pose is registered with the map.
We show, in Figures~\ref{fig:teaser} and~\ref{fig:eval_localization_3_scenes}, how our method can be used for relatively complex scenarios, where the scene is mainly seen from one side, at a constant distance, during mapping and then localization is performed from the other side at varying distances.

\subsubsection{SLAM Resume}

\begin{figure*}[ht]
    \centering
    \includegraphics[width=\linewidth]{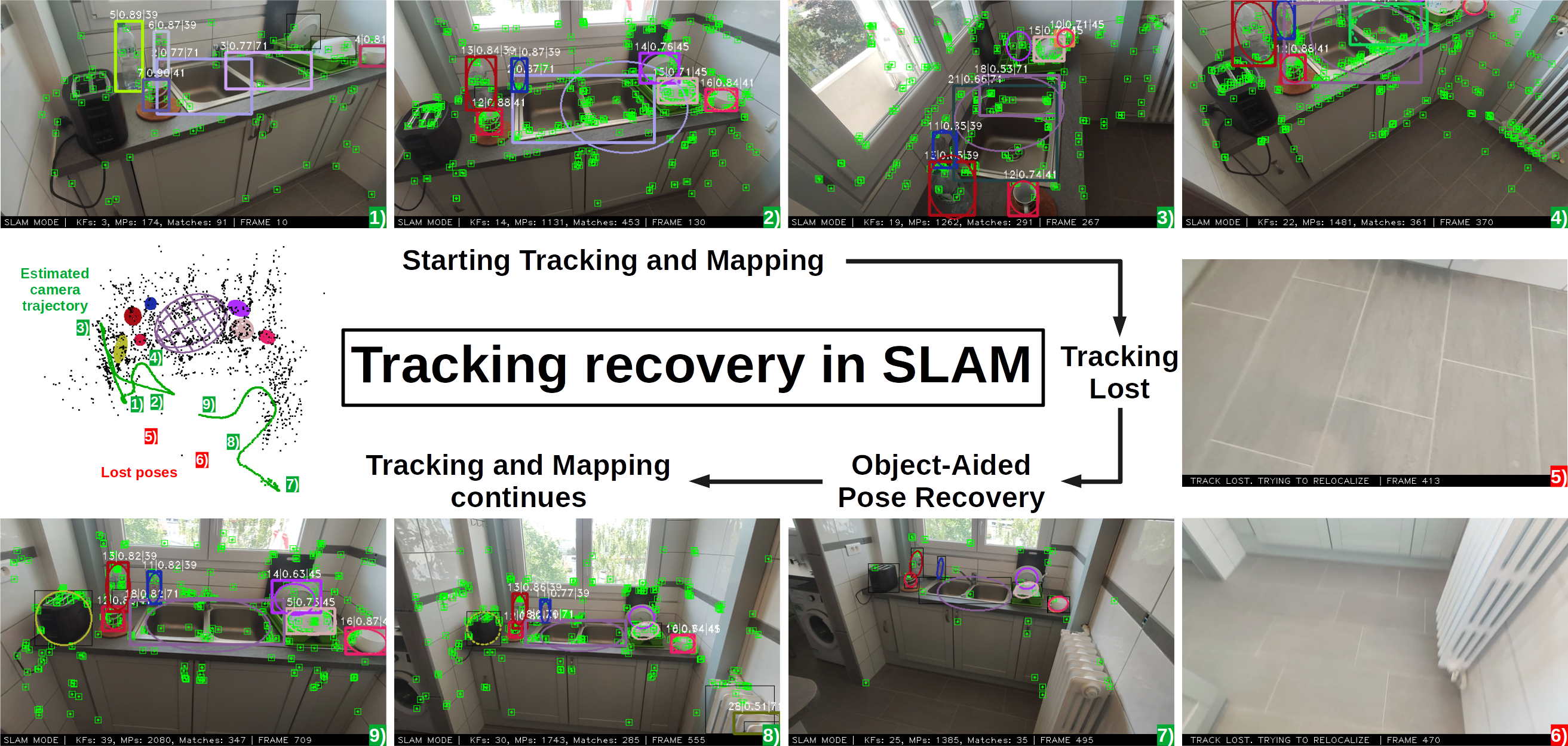}
    \caption{Camera tracking recovery with OA-SLAM. The numbers displayed above the object detections in the images are respectively the id of their associated object, their detection score and their class.}
    \label{fig:eval_recovery_sink}
\end{figure*}

A typical scenario is presented in Figure~\ref{fig:eval_recovery_sink}, which demonstrates the stronger recovery capacity of our SLAM system:
(1-4) The system initially tracks the camera in 3D and builds a map of points and objects. (5-6) The tracking gets lost due to an abrupt camera motion (in this experiment the camera sees only the floor). (7-9) When the reconstructed scene is visible again, the relocalization module estimates the camera pose from objects, establishes point matches and enables the tracking and mapping to continue.
In such a scenario, our object-based approach significantly extends the range of viewpoints from where SLAM recovery is possible, which is limited with the classical point-based method.

In both scenarios, ORB-SLAM2 fails to reinitialize or recover camera tracking during a large number of frames. This is illustrated in the accompanying video.






\subsection{By-Part Modeling}

\begin{figure}[H]
    \centering
    \includegraphics[width=\linewidth]{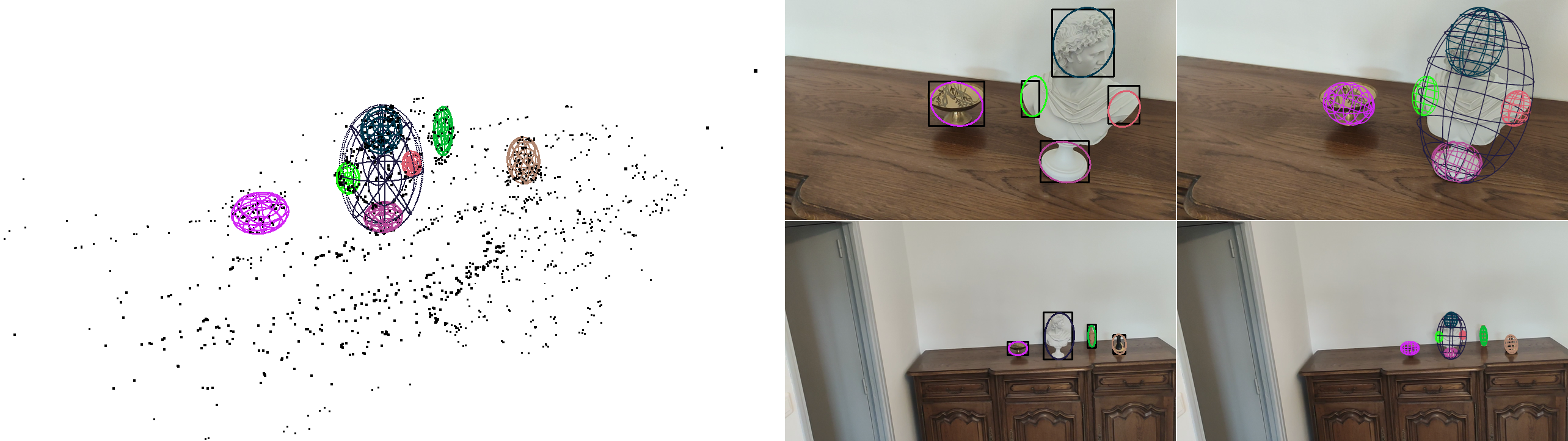}
    \caption{Illustration of by-part modeling. Left: reconstructed map. Right: localized images using complete objects (bottom row) or parts (top row).}
    \label{fig:by_part_modeling}
\end{figure}



Depending on the context, it can be useful to adapt the level of details of the scene modeling (full object vs. parts). When the camera sees the scene from relatively far, objects appear small in the image but are generally in a sufficient number to allow relocalization, which requires at least three objects. In such cases, using global ellipsoidal models is sufficient. However, when the camera gets closer, only one or two objects may be visible and a by-part modeling becomes particularly desirable to increase the number of potential anchors. It is also interesting to better handle objects which are partially occluded.

Our method can handle this in a very flexible way, as the detector network can be fine-tuned to detect distinguishable parts of an object, only requiring some manual annotations in a few images.
Figure~\ref{fig:by_part_modeling} shows the results of our method, after fine-tuning YOLO to detect parts of the statue (head, shoulders and bottom). In the case of a close camera (top row), these parts are used for relocalization, whereas only full object detections are used when the camera is far from the scene (bottom row).




\subsection{Discussion and Conclusion}




In this paper we proposed to integrate objects to point-based monocular SLAM and use them as higher-level landmarks to improve its relocalization ability. Our system leverages existing object detection networks and is able to build a lightweight object map, on the fly.

Comparing our method to the state-of-the-art ORB-SLAM2 system, we show that, thanks to objects, our system is able to relocalize from a significantly larger variety of viewpoints. We demonstrate through experiments that this improved relocalization can be used for initializing 3D tracking on an augmented map or for SLAM tracking recovery after getting lost in the context of AR application. The coarse ellipsoidal models used for objects allow us to reach our objectives of deployment in an unseen world with minimal effort.

While we demonstrated the efficiency of our system in our experiments, it has also some limitations. First, our relocalization approach requires that, at least, three objects present in the map are detected in the query image. This requirement can limit its efficiency, especially in case of a very tight field-of-view, but it can be reduced using a by-part modeling of objects. Secondly, dynamic objects in the scene are not handled by our system. For that, a combination of our object-based reasoning with static/dynamic identification can be considered in future works.





\bibliographystyle{abbrv-doi}

\bibliography{output}
\end{document}